\documentclass[sigconf]{acmart}

    \usepackage{tikz}
    \usepackage{makecell}
    \tikzstyle{decision} = [diamond, draw, 
        text width=5em, node distance=2.5cm, text centered]
    \tikzstyle{block} = [rectangle, draw, 
        text width=7em, text centered, rounded corners, minimum height=4em, node distance = 2.8cm]
    \tikzstyle{line} = [draw, -latex']
    \tikzstyle{cloud} = [draw, ellipse, node distance=2.5cm,
        minimum height=2em]
    \tikzstyle{start} = [draw, circle, node distance=2.5cm, minimum height = 1em]
    \tikzstyle{end} = [draw, circle, node distance=2cm, minimum height = 1em, line width=0.5mm]
    \usetikzlibrary{decorations.pathreplacing}
    \usetikzlibrary{positioning}
    \usetikzlibrary{shapes,arrows,positioning,fit,chains}
    \edef\sizetape{0.5cm}
    \tikzstyle{tmtape}=[draw,minimum size=\sizetape]
    \tikzstyle{tmtapeblack}=[draw,minimum size=\sizetape, fill=black,text=white]
    \usepackage{subcaption}
    \usepackage{amsthm}
    \newtheorem{hypothesis}{Hypothesis}
    \usepackage{soul}

\AtBeginDocument{%
  \providecommand\BibTeX{{%
    \normalfont B\kern-0.5em{\scshape i\kern-0.25em b}\kern-0.8em\TeX}}}

\setcopyright{none}
\renewcommand\footnotetextcopyrightpermission[1]{}
\makeatletter
\renewcommand\@formatdoi[1]{\ignorespaces}
\makeatother

\begin{document}

\title{A Human-Grounded Evaluation of SHAP for Alert Processing}

\author{Hilde J.P. Weerts}
\email{h.j.p.weerts@student.tue.nl}
\affiliation{%
  \institution{Eindhoven University of Technology}
  \city{Eindhoven}
  \country{The Netherlands}
}

\author{Werner van Ipenburg}
\email{werner.van.ipenburg@rabobank.nl}
\affiliation{%
  \institution{Rabobank}
  \city{Zeist}
  \country{The Netherlands}}

\author{Mykola Pechenizkiy}
\email{m.pechnizkiy@tue.nl}
\affiliation{%
  \institution{Eindhoven University of Technology}
  \city{Eindhoven}
  \country{The Netherlands}
}

\renewcommand{\shortauthors}{Weerts, et al.}

\begin{abstract} 
In the past years, many new explanation methods have been proposed to achieve interpretability of machine learning predictions. However, the utility of these methods in practical applications has not been researched extensively. In this paper we present the results of a human-grounded evaluation of SHAP, an explanation method that has been well-received in the XAI and related communities. In particular, we study whether this local model-agnostic explanation method can be useful for real human domain experts to assess the correctness of positive predictions, i.e.\ \textit{alerts} generated by a classifier. We performed experimentation with three different groups of participants (159 in total), who had basic knowledge of explainable machine learning. We performed a qualitative analysis of recorded reflections of experiment participants performing alert processing with and without SHAP information. The results suggest that the SHAP explanations do impact the decision-making process, although the model's confidence score remains to be a leading source of evidence. We statistically test whether there is a significant difference in task utility metrics between tasks for which an explanation was available and tasks in which it was not provided. As opposed to common intuitions, we did not find a significant difference in alert processing performance when a SHAP explanation is available compared to when it is not. 

\end{abstract}



\maketitle

\section{Introduction}
Complex, black-box machine learning algorithms are increasingly applied in fields ranging from medicine to finance. In many contexts, the predictions are input for human decision-makers. Consequently, it is thought to be useful, if not crucial, to understand why a machine learning model made a prediction. However, as the complexity of the models increase, it becomes more difficult for humans to understand their behavior. To tackle this issue, recent efforts in explainable artificial intelligence (XAI) have resulted in many new explanation methods. However, the utility of these approaches in practical scenarios has not been researched extensively. Often, claims about interpretability and utility are based on proxies that have not been evaluated with real humans \citep{DoshiVelez2017, Lage2019}. As explanations need to be interpreted by real humans, a strong but solely theoretical foundation is no guarantee for utility.

In this work, we present a human-grounded evaluation to determine the utility of Shapley Additive Explanations (SHAP) for domain experts who assess the correctness of predictions, such as in medical diagnosis and fraud detection. In particular, we consider the utility for assessment of positive predictions, which we refer to as \textit{alert processing}. SHAP is a state-of-the-art feature contribution method for explaining individual predictions~\citep{Lundberg2017, Strumbelj2014}. To determine the utility of SHAP, we perform two experiments in which real humans perform simplified alert processing tasks while alternately being provided with SHAP explanations. 

\subsubsection*{Methods.} Real humans performed simplified alert processing tasks, with
and without an explanation of the model's prediction. Our approach is two-fold: (1) we statistically test whether there is a significant difference in task utility metrics between tasks for which an explanation was available and tasks in which it was not provided, and (2) we analyze the participants' written reasoning to determine the impact of different sources of evidence on the decision-making process, including the explanation.

\subsubsection*{Main findings.}
In contrast to common assumptions, we did not find a significant difference in alert processing performance between tasks for which a SHAP explanation was shown and tasks for which it was not shown. Our results suggest that possibly SHAP explanations alone are not that useful for alert processing. On the other hand, our qualitative analysis of the participants' reasoning during alert processing suggests that SHAP does affect the decision-making process. We speculate that possibly combining SHAP-based explanations with other techniques may provide higher utility for such tasks.

\subsubsection*{Outline.}
The present paper is structured as follows. In Section~\ref{sec:relwork}, we discuss related work on evaluating explanations of predictive modeling. In Section~\ref{sec:usquestions}, we introduce several ways in which SHAP values may improve task utility as well as the corresponding hypotheses we formulated for the user study. Section~\ref{sec:exp1} and \ref{sec:exp2} cover the experiment setup and results of the first and second experiment respectively. In Section~\ref{sec:userconcl} we present our concluding remarks.

\section{Related Work}
\label{sec:relwork}
Calling for more rigorous evaluations of XAI, \citet{DoshiVelez2017} introduce a three-level taxonomy for evaluating explanations: \textit{application-grounded}, \textit{human-grounded}, and \textit{functionally grounded}. \textit{Application-grounded} evaluations consider the evaluation of real applications with expert users. \textit{Human-grounded} evaluations consider real humans performing simplified tasks that either require or can benefit from interpretability. \textit{functionally-grounded} evaluations use formal proxies of interpretability and do not require research with real human users. 

Our study is an example of a human-grounded evaluation. Although only few works address this type of evaluations, we can identify several evaluation procedures that are commonly applied.

\textit{Output verification} can be used to compare the interpretability of models \citep[e.g.][]{Huysmans2011, Lakkaraju2016, Lage2019}. In this evaluation procedure, participants are asked to verify whether an output is consistent with the model. Another common approach for evaluating the interpretability of an explanation is \textit{forward simulation} \citep[e.g.][]{Huysmans2011, Niklas2011, Strumbelj2014, PoursabziSangdeh2018}. In forward simulations, it is determined how well humans can predict behavior of the model, after being exposed to an explanation \citep{Lipton2016}. Additionally, \citet{DoshiVelez2017} suggest that different explanations can be evaluated by means of \textit{binary forced choice}, in which humans review two alternative explanations of the same model and choose the best alternative.

An evaluation approach closely related to the present paper is \textit{identification of incorrect behavior}. For example, \citet{Ribeiro2016} study whether different explanations methods allow users to identify which classifier is likely to generalize to real world context. 
The capability of users to identify particular cases in which the model makes a wrong prediction under conditions with varying global interpretability was evaluated in~\citep{PoursabziSangdeh2018}. In contrast to common assumptions, the authors found that exposing a model's global internals decreased people's ability to detect mistakes for unusual instances. This result shows the importance of user testing for validating intuitions on the utility of XAI. 

SHAP explanations have been previously evaluated with human users in two ways. A forward simulation experiment, in which SHAP explanations significantly increased predictive performance, was reported in \citep{Strumbelj2014}. Other recent studies \citep[e.g.][]{Lundberg2017, Lundberg2018} show that among several explanation techniques, SHAP corresponds best with human intuitions of a simple decision tree model. Although these results provide evidence on the interpretability of SHAP, it does not directly follow that SHAP is useful for alert processing. 


\section{Hypotheses} 
\label{sec:usquestions}
In this section, we formulate research hypotheses on the utility of SHAP. To this end, we identify cases in which SHAP values might affect task performance for alert processing. We measure task performance in terms of \textit{task effectiveness}, \textit{task efficiency}, and \textit{mental efficiency}, each of which could be impacted by providing a user with an explanation.

\textit{Task effectiveness} refers to the extent to which the system helps the user to perform the task more effectively. For alert processing tasks, task effectiveness can be expressed as accuracy: the proportion of tasks in which the participant correctly distinguished between true positives and false positives.

Compared to just providing a model's confidence score, SHAP explanations could increase task effectiveness by increasing the ability of the user to assess model's credibility. For example, if a certain feature contributes substantially to the model's belief that an instance is positive, but the user assesses this reasoning as counter-intuitive, the user will be more likely to question the model's prediction. Contrarily but equally useful, the SHAP explanation may point the domain expert towards important feature values they would not have considered without being exposed to the explanation.

\begin{hypothesis}
\label{hyp:teffe}
SHAP explanations increase task effectiveness of alert processing compared to the model's confidence scores alone, depending on the reasonableness of the explanation.
\end{hypothesis}

\textit{Task efficiency} refers to the extent to which the system helps the user to perform a task more efficiently, which we express as time spent on the task. Because SHAP explanations reveal which feature values are relevant for the model's decision, they can be used to determine whether the model's explanation is reasonable given domain knowledge. If it is, the domain expert might be able to process the instance more quickly.

\begin{hypothesis}
\label{hyp:teffi}
If the number of features of an instance is sufficiently large, SHAP explanations increase task efficiency invested in alert processing compared to the model's confidence scores alone depending on the reasonableness of the explanation.
\end{hypothesis}

Lastly, \textit{mental efficiency} refers to the required mental resources to perform a task. Mental efficiency can be measured as self-reported mental effort \citep{Paas1992}. According to cognitive load theory, the working memory has a limited capacity. For low-dimensional instances, SHAP explanations are unlikely to improve mental efficiency. Given the increase in information, they may even increase mental effort. On the other hand, a complex instance with many (tabular) features is unlikely to fit into working memory. In such cases, SHAP explanations may help the domain expert to focus on a subset of features that do fit into working memory, unless the explanation is unreasonable. 

\begin{hypothesis}
\label{hyp:me}
If the number of features of an instance is sufficiently large, SHAP explanations increase mental efficiency in alert processing compared to the model's confidence scores alone, depending on the reasonableness of the explanation.
\end{hypothesis}

We test our hypotheses by means of two user experiments in which participants perform alert processing tasks while alternately being provided with SHAP explanations. In the first experiment, we measure the added value of SHAP explanations for each participant compared to the prediction probability alone. In the second experiment, we first explicitly quantify to what extent SHAP explanations agree with human intuitions. Subsequently, we measure the difference in task performance between participants who are provided with an explanation and those who are not. Moreover, we measure whether the difference depends on the extent to which the explanation aligns with human intuition.

\section{Experiment 1: Within-Subject Design}

\label{sec:exp1}
The first experiment is designed to measure the added value of SHAP explanations, compared to the model's confidence score alone. In order to mitigate user-specific effects, we adhere a within-subject experiment design. 

\subsection{Experiment Procedure}
The first experiment consists of alert processing tasks followed by a written reflection. The alert processing tasks are performed in two rounds. The first round is designed for measuring mental efficiency, the second round for measuring task effectiveness.

\subsubsection*{Alert Processing Tasks}
\begin{figure}[ht]
\centering
\includegraphics[width=0.4\textwidth]{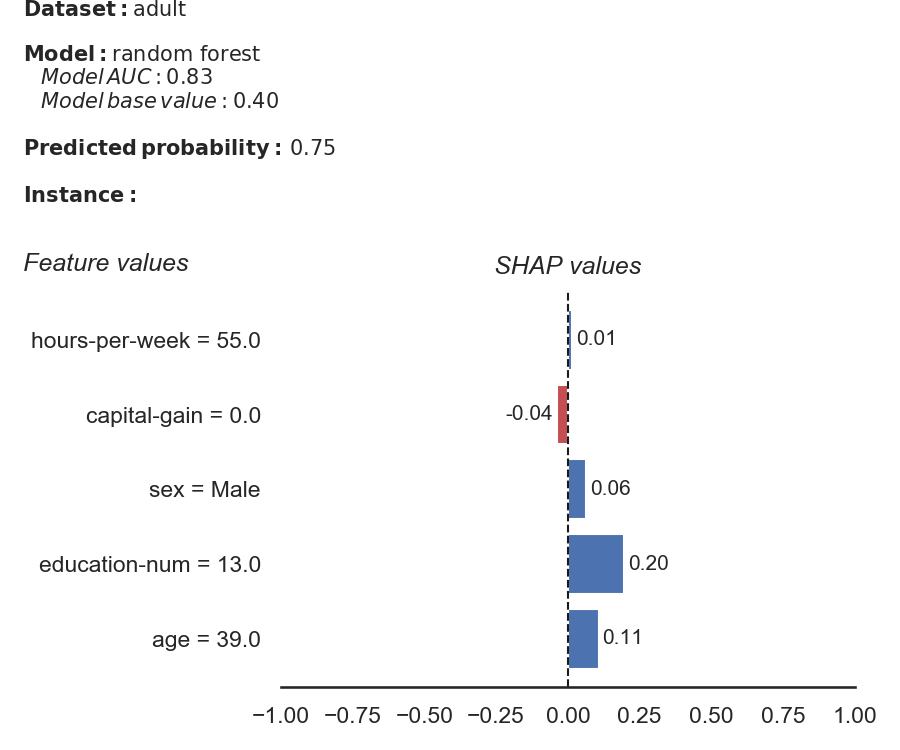}
\caption{Example of an alert processing task in \textit{SHAP} condition. In the \textit{NoSHAP} condition, only the left part of the figure is shown.}
\label{fig:shapexamples}
\end{figure}

In each alert processing task, the participant is provided with an instance the classifier classified as positive. The participant is asked to predict the true class label and how much mental effort they invested to get to their answer. Each task belongs either to the \textit{SHAP} or \textit{NoSHAP} condition. In each task, participants are provided with the model's average confidence score (i.e. base value), the confidence score for the current instance, and the feature values of the instance. In the \textit{SHAP} condition, the participants are also provided with a SHAP explanation (see Figure~\ref{fig:shapexamples}).

\begin{figure} [ht]
    \small
    \centering
    \begin{tikzpicture}
        \begin{scope}[start chain=1 going right,node distance=1mm]
            \node [on chain=1,tmtapeblack] {\textbf{A1}};
            \node [on chain=1,tmtapeblack] {\textbf{A2}};
            \node [on chain=1,tmtapeblack] {\textbf{A3}};
            \node [on chain=1,tmtapeblack] {\textbf{A4}};
            \node [on chain=1,tmtapeblack] {\textbf{A5}};
            \node [on chain=1,tmtape] {{B1}};
            \node [on chain=1,tmtape] {{B2}};
            \node [on chain=1,tmtape] {{B3}};
            \node [on chain=1,tmtape] {{B4}};
            \node [on chain=1,tmtape] {{B5}};
        \end{scope}
    \end{tikzpicture}
    \caption{Setup of \textit{mental efficiency} in Experiment 1. The letter ($A$ or $B$) indicates the instance set, the number (1,2,3,4,5) the instance in the set. The color of the box indicates whether SHAP values are provided (white) or not (black).}
    \label{fig:procexp1me}
\end{figure}
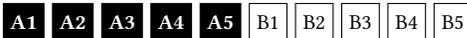

\emph{Round 1: Mental Efficiency.} Participants are provided with two sets of five instances, $A$ and $B$. Each instance in set $A$ is in the \textit{NoSHAP} condition whereas each instance in set $B$ is in the \textit{SHAP} condition. The two sets are shown in order (see Figure~\ref{fig:procexp1me}). 

\begin{figure} [ht]
    \small
    \centering
    \begin{tikzpicture}
        \begin{scope}[start chain=1 going right,node distance=1mm]
            \node [on chain=1,tmtapeblack] {\textbf{1}};
            \node [on chain=1,tmtape] {1};
            \node [on chain=1,tmtapeblack] {\textbf{2}};
            \node [on chain=1,tmtape] {2};
            \node [on chain=1,draw=none] {$\ldots$};
            \node [on chain=1,tmtape] {9};
            \node [on chain=1,tmtapeblack] {\textbf{9}};
            \node [on chain=1,tmtape] {10};
            \node [on chain=1,tmtapeblack] {\textbf{10}};
        \end{scope}
    \end{tikzpicture}
    \caption{Setup of \textit{task effectiveness} in Experiment 1. The number indicates the instance. The color of the box indicates whether SHAP values are provided (white) or not (black).}
    \label{fig:procexp1te}
\end{figure}
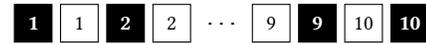

\emph{Round 2: Task Effectiveness.} Participants are provided with one set of ten instances the model predicted to be positives. Each instance is shown twice. The first time, an instance is shown in \textit{NoSHAP} condition, the second time in the \textit{SHAP} condition (see Figure~\ref{fig:procexp1te}). In this setup, any improvement or decrease in task performance will be due to additional information, which means that we can account for the difficulty of the instances. Note that this setup is not suitable for measuring the difference in mental effort, because the same instance is shown twice.

\subsubsection*{Participants' Written Reflections}
After performing the alert processing tasks, participants discuss their results and experiences in small groups. In their written reflections, the groups discuss a.o. whether and how it would have been possible to distinguish between false positives and true positives for each of the ten instances of round 2, \textit{task effectiveness}.

\subsection{Experiment Details}
Choosing an appropriate classification task for this user experiment is not trivial. On the one hand, the classification task should be non-trivial for humans. On the other hand, participants should have some domain knowledge about the data set to be able to argue about the reasonableness of an explanation. In the first experiment, we use the well-known \textit{Adult} data set from the UCI repository \citep{Dheeru2017} which we retrieved from OpenML \citep{OpenML2013}. The related classification task is to predict whether the income of a person exceeds \$50,000 per year based on census data. We select five features and train a random forest classifier using the implementation of scikit-learn \citep{sklearn}. SHAP values were computed using the exact TreeSHAP algorithm proposed and implemented by \citet{Lundberg2018}.

A total of 102 students enrolled in an undergraduate introductory machine learning course participated in the first experiment. 

\subsection{Results of Experiment 1}
To account for multiple testing and retain a family-wise error rate of $\alpha=0.05$, we apply the Bonferroni correction. Accordingly, for each of the tests in the present work, we use a significance level of $\alpha=0.005$.

\subsubsection*{\textbf{Analysis of Task Effectiveness (Hypothesis~\ref{hyp:teffe})}}
\paragraph{Method}
We measure task effectiveness through the proportion of correctly identified false positives and true positives. We test for a difference in the participants' accuracy before and after SHAP values are shown using McNemar's test. A post-hoc two one-sided equivalence test (TOST) procedure is used to assert whether the observed accuracy is equivalent \citep{Lu1995}. We use an equivalence interval of $[-0.05, 0.05]$, i.e. the accuracy of participants within the two conditions is considered equivalent if the difference in accuracy is smaller than 0.05.

\paragraph{Results}
We fail to reject the null hypothesis at a significance level of $\alpha=0.005$ ($\chi^2$(1, N=978) = 0.890, $p = 0.346$). Hence, it can be concluded that the difference in proportion of correct answers between \textit{SHAP} (M=0.61, SD=0.49) and \textit{NoSHAP} (M=0.59, SD=0.49) was not statistically significant. Moreover, the null hypothesis of the post-hoc equivalence test is rejected at $\alpha=0.005$ ($z = -3.20, p_{l}=<.001$, $z = 5.27, p_{u}=<.001$). Hence, we can conclude that there did not exist a meaningful difference in accuracy between the two conditions. 

\subsubsection*{\textbf{Analysis of Mental Efficiency (Hypothesis~\ref{hyp:me})}}
\label{sec:mentaleff}

\paragraph{Method}
Mental efficiency is measured through self-reported mental effort, using the 9-point Likert-scale introduced by \citet{Paas1992}. As some instances may require much more mental effort than others, regardless of the \textit{SHAP} condition, we first perform a one-way ANOVA to determine whether mental effort invested in particular tasks was significantly different from other tasks within either the \textit{SHAP} or \textit{NoSHAP} condition. These tasks are excluded from the remainder of the analysis. Subsequently, we test for difference in average mental effort spent in \textit{SHAP} and \textit{NoSHAP} by means of a two-sided paired t-test. A post-hoc TOST procedure using a one-sided paired t-test is used to assert whether the average mental effort in both conditions is equivalent, considering equivalence interval $[-0.5, 0.5]$.



\paragraph{Results}
For samples related to the \textit{NoSHAP} condition, none of the tasks had a significantly higher mental effort ($F(4, 404) = 1.89$, $p = 0.11$). For the \textit{SHAP} condition, we do find a significant effect ($F(4, 404)= 19.02$,  $p = 0.00$) and a multiple-comparison post-hoc analysis revealed that tasks 2 and 4 required significantly less mental effort than the other questions in the \textit{SHAP} condition. Hence, data points related to these tasks are not considered in the paired t-test.

After asserting that the assumptions of normally distributed differences and the absence of outliers are met, we fail to reject the null hypothesis at $\alpha = 0.005$ ($t(100)=-0.66$, $p=0.51$), which means that the difference in average mental effort in \textit{SHAP} (M=4.81, SD=1.37, SE=0.14) and \textit{NoSHAP} (M=4.74, SD=1.25, SE=-0.12) was not statistically significant. Moreover, the null hypothesis of the equivalence test is rejected ($t_{l}(100) = -4.37, p_{l}=<.001$, $t_{u}(100) = 5.68, p_{u}=<.001$). Hence, we can conclude that there does not exist a meaningful difference in invested mental effort.

\subsubsection*{\textbf{{Analysis of Recorded Participants' Reflections}}}
\paragraph{Method}
A content analysis of the written reflections is performed by means of the grounded theory approach. Each of the reflection reports is coded with regard to the pieces of evidence that were used to make a decision about the true class of each of the instances.

\paragraph{Results}
In total, 22 reports are analyzed. Ten types of evidence are identified, which can be further categorized into four main categories: SHAP values, feature values, the model's confidence score, and similar instances. 

The primary source of evidence described in the written reflections is the instance itself, i.e. its feature values. After feature values, the model's confidence score is mentioned most often. SHAP explanations are used in three different ways. If an explanation is intuitive, participants see this as evidence of the correctness of the prediction. If an explanation is counter-intuitive, participants typically "adjust" the confidence score accordingly. For example, one of the groups argues that: ``\textit{the probability might be on the lower side but the SHAP values show that it is mainly brought down by having a positive capital gain which is quite counter-intuitive.}'' In rare cases, participants change their initial beliefs based on the SHAP explanation. Finally, the true class of a similar instance is sometimes mentioned as evidence.


We would like to stress that the written reflections were performed in hindsight. Hence, the results do not necessarily reflect the participant's behavior \textit{during} the alert processing tasks.

\subsection{Conclusion from Experiment 1}
There was no significant nor a practically relevant difference (larger than 0.05) in task accuracy before SHAP values were shown and after they were shown. Additionally, there was no significant nor meaningful difference (larger than 0.5) in average self-reported mental effort between instances for which SHAP values were provided and instances where SHAP values were not provided.

From the written reflections, we can conclude that apart from the instance's feature values, the leading source of evidence of the true class of an instance was the model's confidence score. This can be alarming, because raw confidence scores are often poorly calibrated; i.e., the predicted probabilities often doe not correspond to the true frequencies \citep{Kuleshov2015}. Consequently, confidence scores may be misleading for domain experts.



\section{Experiment 2: Crossover Design}
\label{sec:exp2}
In the second experiment, we measure the difference in task utility metrics when SHAP values are shown compared to when they are not shown. Recall that Hypotheses~\ref{hyp:teffe}, \ref{hyp:teffi}, and \ref{hyp:me} include a notion of \textit{reasonableness} of the SHAP explanation. In order to quantify the extent to which a SHAP explanation aligns with human intuition, the second experiment is preceded by a pretest experiment in which we measure human-assigned feature value contributions. The experiment setup is adapted from a within-subject to a crossover design.


\subsection{Experiment Procedure}
The experiment setup of the second experiment includes both a pretest experiment and a main experiment.

\subsubsection*{Pretest Experiment}
In order to quantify to what extent SHAP explanations align with human intuitions, we ask participants to assign contributions to feature values of in total 20 instances. For each instance, participants are asked to explicitly indicate to what extent they believe a particular feature value would make it more unlikely, more likely, or would have no impact on the probability of belonging to the positive class. The participants are randomly assigned to two groups, group 1 and group 2. 10 of the instances are evaluated by group 1, the other 10 by group 2. After the data collection, the human-assigned contributions are compared to the corresponding SHAP explanations.
 

\subsubsection*{Main Experiment}
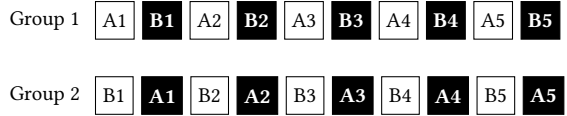
\begin{figure} [ht]
    \small
    \centering
    \begin{tikzpicture}
        \begin{scope}[start chain=1 going right,node distance=1mm]
            \node [on chain=1,tmtape,draw=none] {Group 1};
            \node [on chain=1,tmtape] {A1};
            \node [on chain=1,tmtapeblack] {\textbf{B1}};
            \node [on chain=1,tmtape] {A2};
            \node [on chain=1,tmtapeblack] {\textbf{B2}};
            \node [on chain=1,tmtape] {A3};
            \node [on chain=1,tmtapeblack] {\textbf{B3}};
            \node [on chain=1,tmtape] {A4};
            \node [on chain=1,tmtapeblack] {\textbf{B4}};
            \node [on chain=1,tmtape] {A5};
            \node [on chain=1,tmtapeblack] {\textbf{B5}};
        \end{scope}
        
        \begin{scope}[start chain=1 going right,node distance=1mm, yshift=-1cm]
            \node [on chain=1,tmtape,draw=none] {Group 2};
            \node [on chain=1,tmtape] {B1};
            \node [on chain=1,tmtapeblack] {\textbf{A1}};
            \node [on chain=1,tmtape] {B2};
            \node [on chain=1,tmtapeblack] {\textbf{A2}};
            \node [on chain=1,tmtape] {B3};
            \node [on chain=1,tmtapeblack] {\textbf{A3}};
            \node [on chain=1,tmtape] {B4};
            \node [on chain=1,tmtapeblack] {\textbf{A4}};
            \node [on chain=1,tmtape] {B5};
            \node [on chain=1,tmtapeblack] {\textbf{A5}};
        \end{scope}
    
    \end{tikzpicture}
    \caption{Setup of Experiment 2. The letter (A or B) indicates the instance set, the number (1,2,3,4,5) the instance in the set. The color of the box indicates whether SHAP values are provided (white) or not (black).}
    \label{fig:procexp2}
\end{figure}

In the main experiment, we adhere a cross-over design (see Figure~\ref{fig:procexp2}). Each participant is randomly assigned to either group 1 or group 2. Two sets of instances are considered in the alert processing tasks, set $A$ and set $B$. Group 1 will view instance set $A$ in \textit{SHAP} condition and set $B$ in \textit{NoSHAP} condition. Conversely, Group 2 will see set $A$ in \textit{NoSHAP} condition and set $B$ in \textit{SHAP} condition. In addition to the questions asked in the previous experiment, the participants are asked to provide their reasoning directly after each task; i.e. why they believe a particular instance is a false positive or true positive.

The crossover design has several advantages over the within-subject design of the previous experiment. First of all, in the previous experiment setup, participants had the option to change their answer after being exposed to a SHAP explanation. In the current setup, all information is always shown at once, which better resembles alert processing in a decision support scenario. Second, the new setup allows us to measure all hypotheses in the same experiment. 

\subsection{Experiment Details}

In the second experiment, the UCI \textit{Students Academic Performance} data set is used \citep{Hussain2018}. This data set contains student performance for mathematics in secondary education of two Portuguese schools. The associated classification task is to predict student's grades in mathematics based on a number of features including e.g. \textit{age} and \textit{number of previous failures}. We convert the regression task to a classification task, based on the minimum grade required to pass a course and retain the 13 most predictive features. Compared to the first experiment, the number of features is increased from five to thirteen. Similar to the previous experiment, we have split the data set into a training and test set and trained a random forest classifier.

A total of 20 undergraduate and graduate computer science or industrial engineering students participated in the pretest experiment. The main experiment was executed twice. In total, 57 people participated in the main experiment, consisting mainly of graduate and PhD students majoring in computer science or data science and having basic knowledge of XAI and SHAP.

\subsection{Results of Experiment 2}

\subsubsection*{\textbf{{Analysis of Agreement Between SHAP and Human Intuition (Pretest Experiment)}}}
\paragraph{Method} For each instance, we quantify the agreement between human-assigned contributions and SHAP values as the average correlation. As SHAP explanations typically contain only a few large values and many smaller ones, it is desirable that higher weight is given to the top and bottom ranks. Hence we use a weighted signed rank correlation.



\paragraph{Results}
The average SHAP agreement differs across instances but is typically not much lower than 0. 
For most instances, the model made a correct prediction and the corresponding SHAP explanations agree strongly with human intuitions ($\bar{R}_i > 0.20)$.

\subsubsection*{\textbf{{Analysis of Task Effectiveness (Hypothesis~\ref{hyp:teffe}).}}}

\paragraph{Method}
Given the crossover design, we require a more sophisticated statistical test than in the previous experiment. For Hypothesis~\ref{hyp:teffe}, we use a generalized linear mixed model (GLMM) with a logit link function. We include a fixed effect for \textit{SHAP} condition and agreement with human intuition. Following our hypothesis, we add an interaction effect between condition and agreement, and random effects for the participant and the alert processing task.

\paragraph{Results}
Neither the coefficient corresponding to SHAP (M=0.12, 95\% CI = [-0.71, 0.96], p = 0.76), agreement with human intuition (M=$-7.17$, 95\% CI=$[-13.47, -0.87]$, $p = 0.026$), nor their interaction effect (M=$0.70$, 95\% CI=$[-1.98, 3.38]$, $p = 0.61$) were statistically different from zero.

\subsubsection*{\textbf{Analysis of Task Efficiency (Hypothesis~\ref{hyp:teffi})}}
\paragraph{Method}
A linear mixed effects model is used to test for differences in speed. Again, we include the effect of \textit{SHAP} condition, agreement with human intuition, and the interaction effect between \textit{SHAP} and \textit{agreement}.

\paragraph{Results}
Even after data transformations, normality of the residuals could not be assumed. Hence, no conclusions can be made with regard to task efficiency.

\subsubsection*{\textbf{{Analysis of Mental Efficiency (Hypothesis~\ref{hyp:me})}}}
\paragraph{Method} 
Mental efficiency is measured as self-reported mental effort. A linear mixed effects model is used, including the same effects as for the previous task performance metrics.

\paragraph{Results}
The assumption of normally distributed residuals is reasonable. The SHAP main effect did not significantly differ from zero at $\alpha=0.005$ (M=0.27, 95\% CI = [-0.11, 0.65], p = 0.167), neither did the coefficient of rank agreement (M=-0.35, 95\% CI = [-1.85, 1.15]) nor the interaction effect between SHAP and agreement (M=-0.94, 95\%  CI=[-1.16, 0.272], p=0.129).

\subsubsection*{\textbf{{Analysis of Recorded Participants' Reasoning}}}
Recall that after each alert processing task, the participants are asked to articulate why they believed a certain instance was a false positive or a true positive.

\paragraph{Method}
After several pre-processing steps including stop word removal and lemmatization, the replies are converted to vector-representation that indicates the presence of each term in each of the replies. For each combination of an instance and condition, the proportion of replies that contains a particular term is computed. Subsequently, for each of the instances, the proportions in the \textit{SHAP} and \textit{NoSHAP} conditions are compared. If the absolute percentage point difference between the two proportions is larger than 0.2, the corresponding replies are further inspected manually.

\paragraph{Results}
In total, 20 terms were further inspected. It became clear that some of the differences were due to different wordings (e.g. \textit{study time} versus \textit{studytime}) and in some cases terms were mentioned because the participants did \textit{not} agree with the SHAP explanation (e.g. ``\textit{I do not take into account that much the failures = 0}"). However, in five of the fourteen instances, the participants' reasoning did seem to be affected by the SHAP explanation (see Table~\ref{tab:replies}).

\begin{table}[h]
    \centering
    \small
    \caption{Proportion of replies in which a feature was discussed in \textit{NoSHAP} and \textit{SHAP} condition.}
    \label{tab:replies}
    \begin{tabular}{clrrr}
        \toprule
         Task ID & Feature & SHAP value & \textit{{NoSHAP}} & \textit{{SHAP}} \\
         \midrule
         A2 & \textit{number of absences} & 0.08 & 2/20 & 11/29 \\
         A4 & \textit{previous failure} & 0.05 & 5/20 & 11/25 \\
         A7 & \textit{higher education} & -0.08 & 1/17 & 8/27 \\
         B3 & \textit{number of absences} & 0.04 & 3/30 & 8/22 \\
         B6 & \textit{paid math classes} & -0.01 & 1/28 & 4/16 \\
         \bottomrule
    \end{tabular}
\end{table}

In most of these cases, feature values of the instance are taken into account more heavily when presented with a relatively large SHAP value for that feature value (the largest absolute SHAP values in this data set typically ranged between 0.07 and 0.11). We have not identified any cases in which feature values that were taken into account by people in the \textit{NoSHAP} condition were not taken into account by participants in the \textit{SHAP} condition.

\subsection{Conclusion from Experiment 2}
No significant differences in task effectiveness, task efficiency, and mental efficiency were measured when SHAP values were shown compared to when they were not available to the participants.

From the analysis of the textual replies, it can be concluded large SHAP values did affect the reasoning applied by our participants. These results suggest that large SHAP values can bring feature values of the instance to attention that are otherwise ignored.

\section{Conclusions}
\label{sec:userconcl}
XAI and related research communities have become productive in developing new interpretable machine learning methods. However, the evaluation of these methods often remains limited. 

The results of the present paper suggest that it is important to perform evaluations with real users, rather than to rely on intuitions about utility. In neither of the two experiments it could be concluded whether SHAP explanations significantly impact task utility measured as task effectiveness and mental efficiency. Therefore, we cannot conclude that SHAP explanations are useful for human experts performing alert processing tasks. The post-hoc equivalence tests of our first experiment show that the failure to reject the null hypothesis was likely due to only a small difference in utility rather than a lack of data. This suggests that SHAP explanations alone are not that useful for alert processing. 

Our analysis of the written reflections of participants of Experiment 1 has shown that, apart from the feature values themselves, the leading source of evidence was the model's confidence score. This is concerning, since confidence scores can be misleading. 

Our textual analysis of the participants' reasoning in Experiment 2 has shown that large SHAP values can bring to attention feature values that are otherwise ignored. This shows that even though we could not identify a significant difference in task utility, the SHAP explanations did have an impact on the participants' decision-making process. 

\subsubsection*{Future Work.}



We intend to pursue this direction further by performing a deeper analysis of the gathered data. It would be interesting to identify subgroups in the data for which the difference between \textit{SHAP} and \textit{NoSHAP} is exceptionally large. These subgroups could be described e.g.\ in terms of instance attributes and participant attributes. Such an approach could result in new hypotheses regarding the effect of local explanations on different aspects of alert processing performance. We intend to adopt an exceptional model mining approach introduced in~\cite{DBLP:conf/pkdd/DuivesteijnFPPW17} to automate this search.

Additionally, we would like to replicate the experiments with a classification task that contains a larger number of features and study how this affects task efficiency and mental efficiency.



\bibliographystyle{ACM-Reference-Format}
\bibliography{references}

\clearpage
\appendix

\end{document}